\documentclass{article}

\usepackage{microtype}
\usepackage{graphicx}
\usepackage{subfigure}
\usepackage{booktabs} 
\usepackage{caption}
\usepackage{hyperref}
\usepackage{pifont}
\usepackage[shortlabels]{enumitem}

\usepackage{hyperref}

\usepackage[accepted]{icml2021}

\icmltitlerunning{On The State of Data In Computer Vision: Human Annotations Remain Indispensable for Developing Deep Learning Models.}

\begin{document}

\twocolumn[
\icmltitle{
           On The State of Data In Computer Vision: Human Annotations Remain Indispensable for Developing Deep Learning Models.}




\begin{icmlauthorlist}
\icmlauthor{Zeyad Emam}{scale,zeyad_school}
\icmlauthor{Andrew Kondrich}{scale}
\icmlauthor{Sasha Harrison}{scale}
\icmlauthor{Felix Lau}{scale}
\icmlauthor{Yushi Wang}{scale}
\icmlauthor{Aerin Kim}{scale}
\icmlauthor{Elliot Branson}{scale}

\end{icmlauthorlist}

\icmlaffiliation{scale}{Scale AI, CA, USA}
\icmlaffiliation{zeyad_school}{Department of Mathematics, University of Maryland, MD, USA}

\icmlcorrespondingauthor{Aerin Kim}{aerin.kim@scale.com}

\icmlkeywords{Machine Learning, ICML}

\vskip 0.3in
]



\printAffiliationsAndNotice{}  

\begin{abstract}

High-quality labeled datasets play a crucial role in fueling the development of machine learning (ML), and in particular the development of deep learning (DL). However, since the emergence of the ImageNet dataset and the AlexNet model in 2012, the size of new open-source labeled vision datasets has remained roughly constant. Consequently, only a minority of publications in the computer vision community tackle supervised learning on datasets that are orders of magnitude larger than Imagenet.  \\
In this paper, we survey computer vision research domains that study the effects of such large datasets on model performance across different vision tasks. We summarize the community's current understanding of those effects, and highlight some open questions related to training with massive datasets. In particular, we tackle: (a) The largest datasets currently used in computer vision research and the interesting takeaways from training on such datasets; (b) The effectiveness of pre-training on large datasets; (c) Recent advancements and hurdles facing synthetic datasets; (d) An overview of double descent and sample non-monotonicity phenomena; and finally, (e) A brief discussion of lifelong/continual learning and how it fares compared to learning from huge labeled datasets in an offline setting. Overall, our findings are that research on optimization for deep learning focuses on perfecting the training routine and thus making DL models less data hungry, while research on synthetic datasets aims to offset the cost of data labeling. However, for the time being, acquiring non-synthetic labeled data remains indispensable to boost performance.
\end{abstract}

\section{Introduction}
\label{introduction}

Deep neural networks (DNN) generalize very well to test data \cite{huang2020understanding}. It has almost been a decade since DNNs became widely adopted and DNNs are now the state-of-the-art algorithm on a wide variety of common tasks ranging from computer vision (CV) to natural language processing, graph analysis, and more; however, we still cannot explain the generalization phenomenon exhibited by DNNs. Over the last decade, the research community focused on improving DNN performance by designing better neural network architectures \cite{elsken2019neural}, training optimization routines \cite{ruder2017overview}, and data pipelines \cite{zhang2018mixup, shorten2019asurvey}. Concurrently, ML community has been working on curating novel benchmark labeled datasets. These benchmark datasets are essential in driving new developments because they allow researchers to quickly compare their new methods against previous work.

DNNs require enormous amounts of data to train effectively and generalize robustly in large scale CV settings; however, compiling such datasets is often cost prohibitive. Nonetheless, in some industries, notably the autonomous vehicle (AV) industry, DNNs are deployed at scale in mission critical scenarios, with models continually improving as they train on a never ending stream of data. For instance, Tesla deploys its vehicles to operate as a fleet. Vehicles in the fleet can collect images and video snippets from their surroundings to send back to centralized servers. At Tesla, the data is manually annotated and added to the training set, which is later used to refine their deep learning models. However, collecting and labeling all data points is prohibitively expensive. Therefore, the fleet uses active learning, a set of machine learning algorithms that help users decide which raw unlabeled data is most informative, and therefore worth labeling. By employing active learning, the fleet is selectively collecting "interesting" videos from its surroundings and collecting them to be labeled and used as training data for improving the ML models\footnote{see Andrej Karpathy's talk on Tesla's Autonomy Day  \href{https://youtu.be/Ucp0TTmvqOE}{https://youtu.be/Ucp0TTmvqOE} at 2:06:00.}. To this end, it is important to understand how modern DNNs behave when training with seemingly infinite datasets. Can neural networks benefit from training on seemingly infinite datasets? Do we have the compute power and model capacity to take full advantage of seemingly infinite data? Are the adverse effects of label noise amplified when the datasets are significantly larger, and if so, how can we measure these effects for different computer vision tasks? How does more data affect an algorithm's fairness?

This survey summarizes the community's current understanding of training with seemingly infinite data.

\section{Pre-training with large amounts of labeled data}
Few works in computer vision have tackled the question of training with seemingly infinite labeled data, and none have studied the effects of label noise in that regime, as the data is usually scraped from the internet. 

It is not straightforward to compare the size of two datasets, especially across different data modalities and ML tasks; however, it is widely believed that the largest public computer vision datasets are almost exclusively image classification datasets. The largest open-source image datasets are the Google Open Images dataset (9M images) \cite{kuznetsova2020open}, the Tencent Ml Images dataset (17.5M) \cite{wu2019tencent}, the YFCC-100M dataset (99.2M images) \cite{joulin2016learning}, and the PlaNet dataset (126M images) \cite{weyand2016planet}. Meanwhile, the largest image datasets to feature in publications to-date are Google's internal JFT-300M dataset \cite{hinton2015distilling, sun2017revisiting, chollet2017xception, dosovitskiy2021an}, and Instagram's hashtag dataset \cite{mahajan2018exploring} which contains 3.5 billion images. 

In this section we discuss the most obvious use-case for large datasets: pre-training. We will only survey works that have used the JFT-300M and the Instagram datasets, as they are by far the largest, then we discuss avenues for generating massive datasets cheaply. Later we survey works on self-supervised learning, and finally, we study the limitations of pre-training and the trade-offs between pre-training and fine-tuning dataset sizes.

\subsection{JFT-300M and Instagram Hashtag Datasets.} \label{sec:jft-300m}
The JFT-300M image classification dataset is an internal dataset used by Google for ML research and applications. The dataset contains 300M images with 375M noisy labels across 18,291 classes - approximately 20\% of the labels are incorrect. The dataset distribution is heavily long-tailed - e.g. 2M images of flowers but more than 3k classes with less than 100 images per class, and more than 2k classes with less than 20 images per class. Here, we discuss the major takeaways from papers \cite{sun2017revisiting, dosovitskiy2021an, brock2021highperformance} that report results on JFT-300M. These works pre-train neural networks on JFT-300M then finetune them on different target datasets across different tasks: image classification on Imagenet \cite{deng2009imagenet}, object detection on COCO \cite{lin2015microsoft} and PASCAL-VOC \cite{everingham2015thepascal}, semantic segmentation on COCO, and human pose estimation on COCO. \\

The Instagram hashtag dataset \cite{mahajan2018exploring, dosovitskiy2021an} is composed of 3.5B images and 17k classes. It is unknown how much label noise the dataset contains. The dataset is publicly visible, such that users can look at the images through Instagram, however it is not possible to scrape the data and use it for research purposes. The papers surveyed here pretrain on Instagram data and finetune on various subsets of ImageNet. 

The major takeaway is simple to state: \textit{if the pre-training task is similar to the target task, then pre-training on massive datasets unequivocally helps}. We list here other takeaways in order of relevance. These takeaways assume the pre-training and target tasks are both natural image classification, if the tasks were different these observations don't necessarily hold - we discuss this case in section \ref{sec:tradeoff-pretrain}.

\setlist{nosep}
\begin{enumerate}[noitemsep]
    \item \textbf{Logarithmic Improvement:} \cite{sun2017revisiting, mahajan2018exploring} have found the network's performance to grow as a logarithmic function of the pre-training dataset size.
    \item \textbf{Model Capacity:} Larger models are crucial to fully capture the performance boost offered by pre-training on massive datasets \cite{sun2017revisiting,dosovitskiy2021an, mahajan2018exploring}. All papers found that pre-training performance appears bottle-necked by model capacity. We will tackle model capacity in more detail when discussing the double descent phenomenon in section \ref{sec:double-descent}. 
    \item \textbf{Fine-tuning Hyperparameters}: Fine-tuning hyperparameters are drastically different from the parameters used during pre-training. \cite{mahajan2018exploring} found that Instagram data requires fine-tuning learning rates that are up to an order of magnitude lower than pre-training.
\end{enumerate} 
In the case of \cite{sun2017revisiting}, pre-training on JFT-300M took 2 months across 50 Nvidia K80 GPUs. Meanwhile \cite{mahajan2018exploring} managed to pre-train their model on the full Instagram dataset in 22 days across 336 GPUs. It is not possible to conduct a thorough hyperparameter sweep in this regime. Therefore, these efforts present a lower bound on the actual performance gains that can be achieved when training on these massive datasets. Additionally, both papers did not saturate on model capacity.

\subsection{Synthetic datasets}


Large-scale real-world labeled datasets are notoriously challenging to collect. Synthetic data presents an attractive shortcut, enabling organizations to quickly collect large amounts of task-relevant and strongly-labeled data. However, because neural networks are notoriously sensitive to distribution shift, the common wisdom is still to finetune a model trained on synthetic images with real data from the target distribution. In this section, we investigate how combining real and synthetic data impacts performance on a variety of computer vision tasks. 

We note that for some computer vision applications, synthetic data is the only way to gather appropriate labels since human annotation is impossible or time consuming. For example, geometric information such as optical flow and depth can only be reliably generated in simulation \citep{DBLP:journals/corr/McCormacHLD16} and it is challenging to collect noise-free human annotations for robot manipulation \citep{DBLP:journals/corr/abs-1811-02790, DBLP:journals/corr/abs-1911-04052}. We consider here semantic tasks like semantic segmentation or 2D box annotation where human labels are unambiguously correct and can scale to massive dataset sizes.

For self-driving applications, finetuning a model pretrained on synthetic data on a small real dataset can significantly improve performance. In some instances, pretraining on synthetic data is the single most beneficial addition to the training pipeline as it exhibits the highest performance gains. A seminal work in this space, \cite{DBLP:journals/corr/RichterVRK16} demonstrated that a model trained with just 1/3 of the real data along with synthetic renders from a video game engine could outperform training on all the real data on the CamVid semantic segmentation task \cite{10.1016/j.patrec.2008.04.005}. However, the real data was crucial to achieving this performance, accounting for an increase of over 20 percentage points from the zero shot synthetic data baseline. MetaSim \cite{DBLP:journals/corr/abs-2008-09092} learns to generate scenes that match the content found in real scenes. The paper demonstrates significant gains on the Kitti \cite{6248074} validation test from simply finetuning a model trained with MetaSim synthetic scenes on 100 Kitti training samples, reporting an increase on the "moderate" data category from 66.3 AP @ 0.5 IOU to 73.9 on MetaSim data and an increase from 63.7 to 72.7 on baseline synthetic data. In comparison, we only observe an increase from 66.3 to 67 from MetaSim to MetaSim2 \cite{DBLP:journals/corr/abs-1904-11621}, an extension on the method that learns scene structure alongside parameters. For practical consideration, both of these results suggest that collecting a small dataset in the target distribution can be better for downstream performance than iterating on the quality of the synthetic data. 

Methods that use deep generative models to mimic the task distribution exhibit better zero-shot transfer performance compared to graphics-based approaches which are typically used in self-driving tasks. \cite{DBLP:journals/corr/abs-2009-00668} consider generating synthetic volumetric images for medial applications. The authors use a conditional generative adversarial network (GAN) \cite{goodfellow2014generative, DBLP:journals/corr/MirzaO14} to generate shape and material maps which are then fed through a differentiable CT renderer to create a voxel map. Models trained only on GAN-generated images were able to closely match and outperform baselines trained on the full original dataset, even though the GAN model was trained on a small subset of the data. \cite{DBLP:journals/corr/abs-2104-06490} trains a small MLP on StyleGAN \cite{DBLP:journals/corr/abs-1812-04948} features to additionally predict labels for each pixel, where human annotators segment model-generated images. Performance on downstream segmentation tasks similarly shows models trained on these images can consistently outperform "upper bound" models trained on the fully labeled dataset. In both cases, we see that GAN generated synthetic images seem to transfer well to downstream tasks, provided a set of human annotations to learn to imitate.

In conclusion, accessing the desired data distribution tremendously improves transfer performance from synthetic data to the target domain. For models trained on rendered scenes using traditional graphics, finetuning on even a small set of real visual scenes results in significant gains relative to iterating on the network's architecture, optimization routine, or the quality of synthetic data. Approaches to synthetic data that directly mimic the real data distribution have shown great zero shot transfer, but still rely on the initial set of data to train the generative model and are currently limited to certain task types.

\subsection{Self-supervised learning}

Self-supervised learning gives us effective representations and is a good starting point for downstream tasks in an environment where it's nearly impossible to annotate all the data available. However, large-scale labeled data has historically required to train deep learning models in order to achieve better performance in both Natural Language Processing (NLP) and CV domain. Are self-supervision or predictive unsupervised learning beneficial when massive labeled datasets exist for the downstream tasks? \cite{DBLP:journals/corr/abs-1906-12340, DBLP:journals/corr/abs-1902-06162} showed that while self-supervision does not substantially improve accuracy when used in tandem with standard training on annotated datasets, it can improve different aspects of model robustness: noisy labels \cite{DBLP:journals/corr/abs-1805-07836}, adversarial examples \cite{madry2019deep}, and common input corruptions such as fog, snow, and blur \cite{DBLP:journals/corr/abs-1903-12261}. Additionally, self-supervision has also shown its effectiveness in out-of-distribution detection on difficult and near-distribution outliers \cite{DBLP:journals/corr/abs-1906-12340} and has been essential in video tasks where annotation is costly \cite{DBLP:journals/corr/VondrickPT16, DBLP:journals/corr/abs-1806-09594}.

Since the introduction of Word2Vec  \cite{mikolov2013efficient} and Glove \cite{pennington-etal-2014-glove} self-supervised learning has been more extensively used in NLP compared to CV domains. Pretraining embeddings on large unlabeled text and finetuning them on large-scale labeled datasets enabled large performance boosts. To take advantage of these massive unlabeled datasets, researchers developed adequately large models such as BERT  \cite{DBLP:journals/corr/abs-1810-04805}, RoBERTa \cite{liu2019roberta}, and XLM \cite{DBLP:journals/corr/abs-1901-07291}. These models are trained in a predictive fashion, meaning the model is trained to predict the words or tokens that were masked or replaced. Upon obtaining the pretrained embedding, we resume training, this time with supervision, for different types of NLP downstream tasks such as question answering, machine translation, summarization, natural language inference, sentiment analysis, semantic parsing. \cite{lecun2021blog} pointed out that NLP's discreteness is the reason why it can make deep learning prediction problems tractable because, in NLP, there is only a finite number of words to be predicted. This also makes NLP problems great candidates to use predictive architectures. On the other hand, discreteness makes some tasks more difficult. For instance natural language generation usually fails when the input text is modified, even if its meaning is preserved, or when slight perturbations are added to the learned embeddings \cite{DBLP:journals/corr/abs-1711-02173, fedus2018maskgan, DBLP:journals/corr/abs-1805-02917, iyyer-etal-2018-adversarial, dong2021towards}. In contrast, small perturbations are not really perceptible in generated images or audio, which are continuous signals. 

Compared to NLP, the development and the adoption of self-supervision in CV domain have been relatively limited. \cite{dosovitskiy2021an} trained a network to discriminate between surrogate classes which are formed by applying  transformations to a randomly sampled seed image patch. In contrast to supervised training, the resulting feature representation is not class specific; however, this generic representation improves classification results. \cite{DBLP:journals/corr/DoerschGE15} predict the relative position of image patches and use the resulting representation in order to
improve object detection. \cite{DBLP:journals/corr/abs-1803-07728} predict image rotations instead. Other proxy tasks have also been suggested, such as predicting per-pixel color histograms \cite{DBLP:journals/corr/LarssonMS16}, DeepCluster, a clustering method that jointly learns the parameters of a neural network and the cluster assignments of the resulting features \cite{DBLP:journals/corr/abs-1807-05520}, and maximizing mutual information between an input and the output of a deep neural network encoder \cite{hjelm2019learning}. These works focus on the utility of self-supervision for learning without labeled data and do not considered its effect on robustness and uncertainty.

Most recently, vision transformers (ViT) \cite{DBLP:journals/corr/abs-2010-11929} have shown impressive performance across various CV problems while requiring lower computational resources to train. These models are based on a multi-head self-attention architecture that can attend to a sequence of image patches to encode contextual cues. DINO \cite{DBLP:journals/corr/abs-2104-14294} developed self-supervised ViT features which contain explicit information about the semantic segmentation of an image. These features do not emerge as frequently when training ViTs in a fully supervised fashion. \cite{bao2021beit} propose a masked image modeling task, one of the successful NLP pre-training approaches in order to pretrain ViTs. Their method tokenizes the original image into visual tokens then create a masked image modeling task.

Ultimately, self-supervision can dramatically reduce the number of annotations required to train. However, without the paradigm shift from the well-established workflow of ML practitioners – pretraining then fine-tuning, human in the loop learning including the data annotation will remain indispensable to train deep learning models on various machine learning tasks. For the time being, self-supervision is best used in tandem with full-supervision. This can induce strong regularization that improves robustness, while requiring fewer manual annotations to train.

\subsection{Striking a balance: the tradeoff between pre-training and fine-tuning dataset sizes.}\label{sec:tradeoff-pretrain}
 
Conventional wisdom in computer vision up until recently was to pre-train models on a large labeled image classification dataset before fine-tuning on a target dataset tailored for a specific use-case. Recently, however, \cite{he2019rethinking} challenged this notion by showing that pre-training on ImageNet does not boost performance for object detectors on the COCO dataset. So how does performance change as we vary the ratio of pre-training to target dataset sizes? 
This question is particularly relevant when deciding which data to spend labeling effort on. Consider for instance a large company looking to allocate funds for data labeling across multiple teams, with each team working on a slightly different computer vision task. How should the funds be allocated? Is it better to allocate more budget for labeling a large general purpose dataset -with cheaper classification-only labels- that would be shared across teams to pre-train their respective models? In this scenario each team would receive less funding to label their specialized, target datasets, which include more expensive labels - i.e. bounding boxes or segmentation labels. Or would it be better to avoid funding the shared dataset altogether and allocate the entire budget evenly across teams, enabling each team to work with larger specialized datasets?

In \cite{he2019rethinking, ghiasi2018dropblock}, the authors show that pre-training on ImageNet does not help when fine-tuning on a different task, namely object detection on COCO; \cite{poudel2019fast} make a similar remark for image segmentation on CityScapes \cite{cordts2016cityscapes}; finally, and perhaps less surprisingly \cite{raghu2019transfusion} also find ImageNet pre-training ineffective when fine-tuning on medical images. \\
It is true that if a model is initialized from a pre-trained checkpoint and fine-tuned on a target dataset, it will train faster than a model that is randomly initialized. This speed-up is in part what has fueled the pre-training craze. However, \cite{he2019rethinking} show that if the randomly initialized model trains to saturation, then both models exhibit a similar performance. This observation holds true even when the target dataset is significantly smaller than the pre-training dataset. The authors only find significant performance deterioration for the randomly initialized model when using less than 10\% of the COCO dataset\footnote{The authors do not narrow down the exact point of performance deterioration, but find that it is somewhere between 10\% and 3.5\% of the data.}. In this case, fine-tuning outperforms training from scratch.\\
Similarly, in section 3.2 of \cite{mahajan2018exploring}, the authors tried pre-training on Instagram's hashtag dataset (an image classification dataset) and fine-tuning on COCO object detection. They computed the results using the default COCO average precision (AP) metric and AP@50. The former emphasizes exact object localization, while the latter allows loose localization and emphasizes correct object classification. They have found that pre-training provides a significantly larger performance boost with respect to the AP@50 metric over the AP metric. In other words, pre-training helped the model achieve better classification (same task), but not necessarily better localization (different task). \\
Furthermore, even when both the pre-training and target datasets are natural image classification datasets, \cite{mahajan2018exploring} noticed that by pruning the pre-training dataset and only keeping those images with labels matching the target dataset's labels, the model pre-trained on the pruned Instagram dataset outperforms the model pre-trained on the entire Instagram dataset. This is further evidence that aligning the pre-training and target tasks as much as possible is more beneficial than collecting more pre-training data across a larger range of classes. These observations are inline with our discussion on synthetic datasets above. 

There are many open questions in this area, especially when the target dataset is small. However, these results show that for mission critical applications, it is necessary to collect a large amount of data for the target dataset to achieve adequate performance. This is true regardless of the pre-training dataset's size or quality. In this regime, since the target dataset is already large, it follows that pre-training is unnecessary. Finally, following the observation in \cite{he2019rethinking}, we would like to clarify that the results presented here are not an argument against the community's ultimate goal, namely universal representations; in fact, the results laid out here should help guide future dataset curation efforts towards achieving this goal.

\section{Lifelong/Continual Learning and Active Learning.} \label{sec:active-learning}

The computer vision research usually take place in an isolated setting, compared to industry cv applications. The researcher chooses a fixed model architecture and large labeled datasets, trains the model on the training set until it achieves adequate performance on the test data, then deploys the model. There are many underlying assumptions that take place in this \textit{isolated learning} paradigm: 
\setlist{nosep}
\begin{enumerate}[noitemsep]
    \item Closed-world assumption: Performance on the test data is indicative of performance upon deployment. 
    \item Fixed-model assumption: Keeping the model architecture fixed upon deployment will not poorly affect performance.
    \item Data distribution is fixed: The training and test data-sets are drawn from the same \textit{fixed} distribution encountered in real life. There is no need to continually refine the model on newly collected data samples. 
\end{enumerate}
These assumptions limit the use-case of DNNs. On the other hand, humans take a different approach to learning, by continuously accumulating knowledge and adapting to our surroundings. Lifelong machine learning is a paradigm under which ML models can learn in a similar manner as humans. \cite{chen2015lifelong, liu2016lifelong} define a lifelong learning algorithm as an algorithm that has performed a sequence of $N$ learning tasks, $T_1, T_2, \ldots, T_N$ . When faced with the $(N+1)th$ task $T_{N+1}$ with data $D_{N+1}$, the learner can leverage the prior knowledge in its memory to help learn $T_{N+1}$. The memory stores and maintains the knowledge learned and accumulated in the past learning of the $N$ tasks. After learning $T_{N+1}$, the memory is updated with the learned results from $T_{N+1}$. The algorithm should then hopefully be proficient in all tasks $T_1, T_2, \ldots, T_{N+1}$. We purposefully do not define the concept of a "task" as it is very application dependent. In general, a new task can be thought of as a new batch of data with new class labels, or a new domain distribution, or a new output space \cite{delange2021continual}.

Lifelong learning for DNNs assumes infinite training data. A prominent example that combines lifelong learning with DL can be found in the autonomous vehicle industry. Self-driving car companies intend to operate their vehicles as a fleet. Vehicles in a fleet use an object detection DNN to identify moving objects and avoid collisions. This DNN was trained offline on a computing cluster using a large labeled dataset. After training, identical duplicates of this DNN are deployed to every vehicle in the fleet over the internet.\\
Simultaneously, vehicles in the fleet are constantly applying data querying algorithms \footnote{These algorithms constitute a separate branch of ML called active learning.}, if the querying algorithm deems a scene particularly relevant, the vehicle sends the scene back to the cluster to be annotated and added to the training set. The DNN is then re-trained on the newly acquired data and redeployed on the fleet. This cycle allows the fleet to continuously improve its driving performance by refining its model. \\ 
The above example is still closer in nature to the isolated learning paradigm than it is to the general lifelong learning paradigm. For instance, the network does not need any concept of memory since it can retain old training data. In other words, the training data is not treated as a stream where each new batch is discarded after training the model; instead, all previously collected data is stored on the server, and the network can be trained offline on all the available data\footnote{This type of continual learning falls under "replay methods using rehearsal". See section 3.1 of \cite{delange2021continual}.}. More generally, a lifelong learning algorithm should be able to operate even when training data pertaining to previous tasks expires\footnote{see section 7 in \cite{delange2021continual} for a list of characteristics that an "ideal" lifelong learning algorithm should incorporate.}. 

\subsection{Catastrophic Forgetting}
When training data pertaining to previous tasks cannot be re-used, lifelong learning becomes more difficult. DNNs are prone to catastrophic forgetting. This phenomenon is one of the biggest hurdles for achieving lifelong learning across a wide range of tasks. Catastrophic forgetting occurs when a network, previously trained to perform a certain task, is now trained to perform a new task. When training on the new task, the neural network is susceptible to abruptly forget how to perform the original task. In this case, the model must implement some form of (implicit or explicit) memory to avoid forgetting the older tasks. There is a large body of research concerned with designing the memory \cite{delange2021continual}, including, pseudo-rehearsal methods \cite{shin2017continual} where the memory is in the form of a generative model trained on previous tasks, regularization-based methods \cite{kirkpatrick2017overcoming} which focus on finding network parameters that can learn new tasks without forgetting older tasks, and parameter isolation methods \cite{aljundi2017cvpr} which introduce a new set of parameters to the network every time a new task is encountered. \cite{delange2021continual} compare several of these methods by dividing Tiny ImageNet (a subset of 200 classes from Imagenet with each class containing 500 images) into different tasks, each containing a different set of labels, they find that parameter isolation methods perform best, followed by regularization methods. \\ Unfortunately, to the best of our knowledge, lifelong learning research still lags behind conventional offline CV research, and there are no results on lifelong learning in the huge data regime.  

\section{Model Selection} 

In section \ref{sec:jft-300m} we mentioned that a good choice of model capacity is crucial in order to fully exploit massive datasets. In this section we study the effect of model capacity (also known as complexity) on the performance of DNNs trained on massive amounts of data\footnote{We are intentionally using the term "model capacity" loosely as there is not a single agreed upon definition in the literature and a discussion of formal definitions is outside the scope of this paper.}. A large body of works stemming from the ML theory community studies this effect in an attempt to explain the puzzling generalization exhibited by deep neural networks - see section 3 of \cite{huang2020understanding} for a brief survey. However, there still remains a lot of unanswered questions in this area. Here we focus on more empirical observations from the literature regarding the relation between model complexity and training on massive datasets, then we tackle the double descent and sample non-monotonicity phenomena and what they entail in the massive data regime.

The papers surveyed in section \ref{sec:jft-300m}  \cite{mahajan2018exploring, sun2017revisiting, dosovitskiy2021an} found it important to use appropriately large models when training with massive datasets, as small models are susceptible to under-fitting the data. In fact, \cite{mahajan2018exploring} have found that pre-training small models on more and more data can have a negative effect on performance after fine-tuning on the target task. In \cite{hestness2017deep}, the authors built on prior theoretical works, by devising experiments to determine the optimal model capacity as a function of dataset size. In their experiments, the optimal model complexity increases sub-linearly according to a power law with the size of the dataset. As the authors note, their experiments consistently exhibited this power law behavior, suggesting that their line of work could be leveraged to systematize a host of tasks, such as neural architecture design and model size selection.\\

\subsection{Double Descent and Sample Non-Monotonicity}\label{sec:double-descent}
The double descent phenomenon postulates that medium sized models are more susceptible to overfitting than their smaller or larger counterparts. The name stems from plotting the test errors of different ML models trained for a fixed large number of epochs against the number of parameters in each model. As model complexity increases, the test error will start off as a large number, decrease, increase, then finally decrease again. The middle region on the plot where the loss increases as a function of model complexity is called the critical regime. Double descent is a fairly universal phenomenon that occurs across a variety of tasks, architectures, and optimization methods.  \\

Neural networks defy classical statistical learning theory because they do not overfit despite having more parameters than samples in the training set. Double descent was first postulated in generality in \cite{belkin2019reconciling}, who demonstrated it for decision trees, random features, and 2-layer neural networks with $l2$ loss, on a variety of learning tasks including MNIST and CIFAR-10. It has been theoretically analyzed in the tractable setting of linear least squares regression \cite{belkin2019reconciling,hastie2020surprises,bartlett2020benign,mitra2019understanding}. In \cite{nakkiran2019deep}, the authors find that the double descent phenomenon for DNNs is most prominent in settings with added label noise. When using early stopping to train deep neural networks, the double descent phenomenon becomes less pronounced. In other words,\textit{ the set of model complexities that are susceptible to overfitting shrinks, making it highly unlikely to select a poor model.} \\
Early stopping is easy to implement and is already widely adopted by ML practitioners; on the other hand, label noise is a common and difficult problem to eliminate. It is therefore important to carefully select models when training with noisy data as is typically the case in the massive dataset regime. 

In their work, \cite{nakkiran2019deep} also evaluate the test performance as a function of the number of training samples. Sample non-monotonicity is a phenomenon that occurs when comparing two medium sized models trained for a fixed (large) number of epochs on different amounts of data. Surprisingly, the model trained with less data outperforms its counterpart. This observation only holds for medium-sized models, and only after training for a fixed number of epochs. \textit{In the presence of early stopping the authors could not reproduce sample non-monotonicity for any model complexity.}

While there are still many open questions regarding optimal model sizing in small data regimes, we are far from saturating on model capacity in massive data regimes. In general, we conclude that, given sufficient labeled data, with little label noise, larger models trained with early stopping are generally more likely to outperform smaller models.

\section{Huge Datasets and Algorithmic Fairness}

As applications of machine learning grow more widespread, it becomes increasingly important to understand the sociopolitical implications of algorithmic decision making. Numerous studies have shown that across applications, deep learning models have a tendency to reinforce existing inequalities.  For example, authors \cite{buolamwini2018} showed that commercial face recognition algorithms have significantly higher error rates for dark-skinned women than light skinned men, and a 2019 NIST study showed that a majority of facial recognition algorithms misidentify racial minorities  10 to 100 times more often than whites \cite{nist2019}.

Why is racial and gender bias so pervasive in deep learning applications? The majority of evidence suggests that non-diverse training datasets are the largest factor driving racially disparate results. Indeed, many open source facial recognition datasets are biased towards light-skinned faces.  For example, Labeled Faces in the Wild is 83.5\% white, and the IJB-A dataset, which was specifically created to emphasize geographic diversity, draws only 21.4\% of its examples from faces with darker skin tones \cite{buolamwini2018}. Researchers \cite{utkfairface} compared models trained on the UTKFace dataset with models trained on the FairFace dataset (which is designed to be balanced with respect to race), and found that across 3 model architectures, model results were substantially less biased after training FairFace \cite{utkfairface}. These findings underscore the need for better public, large scale datasets labeled with demographic data in order to enable further empirical study of algorithmic bias. The effect of dataset size and dataset balance on algorithmic fairness remains an open question in the literature.

Training datasets collected in the real world contain societal and historical biases\footnote{In this review we've omitted the topic of federated learning. Federated learning is concerned with training using data available only on edge devices, please refer to \cite{kairouz2019advances} for an overview. }. \cite{fairsynthetic} are hopeful about the potential for synthetic data to incorporate fairness constraints.  Using a generative model, they produced a synthetic version of the UCI Adult census dataset which resulted in a more fair classifier \cite{fairsynthetic}. However, whether such methods will prove effective for more complex and unstructured data types such as images remains an open research question.

\section{Label Noise}

The labels of datasets crawled from the Internet are inherently noisy. For example, in the Instagram-1B dataset, hashtags are not necessarily visually relevant. \cite{mahajan2018exploring} found that when 50\% of hashtags are randomly replaced (i.e. noise injection), the accuracy of the target task is reduced by 7.31\% to 14.29\% depending on the number of classes in the target task. This behavior persisted when tested in smaller datasets as well. Meanwhile, \cite{rolnick2017deep} showed that test accuracy above 90 percent is achievable on smaller datasets (e.g. MNIST) even after each clean training example has been diluted with 100 randomly-labeled examples.
These results suggest that we could potentially identify and prune a certain percentage of data with no effect on accuracy. Therefore, when it comes to pretraining, resources are generally better spent on gathering additional labeled data in the downstream target dataset rather than improving the quality of labels in the pretraining data. 

To model systematic label noise (or class-dependent label noise), there exists methods such as loss correction \cite{patrini2017making} or label smoothing \cite{lukasik2020does}. In particular, \cite{patrini2017making} produces a stochastic matrix indicating the probability of one class being mislabeled as another class and shows that, on the Clothing 1M dataset \cite{xiao2015learning}, accuracy is improved by 16.59\%.

\section{Conclusion}
We've surveyed several topics related to training neural networks on seemingly infinite data. Due to budget constraints and the lack of large-scale open-source datasets, only a few members of the ML community are able to conduct experiments at massive scale. However, there are many under-studied areas of research where experiments with relatively smaller datasets (e.g. ImageNet) could yield scientifically interesting results. For instance, how can we achieve better active learning? Not all training examples are created equal, some could be more informative than others, and ideally we should focus our labeling effort on annotating more informative samples. There are other promising research areas which revolve around fairness: How might we use synthetic data to improve deep learning performance relative to quantitative definitions of fairness? How do we ensure models are fair when training on a long tailed distribution? 

Despite years of research, we still cannot clearly pin point why over-parameterized DL models can generalize so well. Solving this mystery will require the community to expand on existing statistical learning theory. This improved theoretical paradigm will pave the way for a myriad of research directions, including fundamental questions such as: how should we design a model's architecture given some data? how well can we predict the model performance, given a specific model and data? And, how much data do we really need to achieve adequate generalization? 

To conclude, training on seemingly infinite data is hard, and we're only beginning to understand the benefits and pitfalls of training in this regime. Training in this regime is costly at every step: collecting and labeling the data is a costly process whether carried out manually, via generative models, or via graphics engines.  Training large enough models requires access to a lot of compute resources, and finally, deploying and maintaining such models at scale can also be very costly, especially if they're continually learning. In the long run, the cost of each step will be reduced. Our point of concern -- the cost of human annotation -- also should be mitigated through large-scale self-supervised learning. However, without the paradigm shift from the well-established workflow of ML practitioners -- pretraining then fine-tuning, human in the loop learning including the data annotation will remain indispensable to train deep learning models on various machine learning tasks.



\bibliography{main}
\bibliographystyle{icml2021}

\end{document}